\documentclass[10pt,twocolumn,letterpaper]{article}

\usepackage{3dv}
\usepackage{times}
\usepackage{epsfig}
\usepackage{graphicx}
\usepackage{amsmath}
\usepackage{amssymb}

\usepackage{mathtools}
\usepackage{amsmath}

\usepackage{amssymb}
\usepackage{amsfonts}
\usepackage{amsopn}
\usepackage{graphicx} 
\usepackage{textcomp}
\usepackage{xfrac}
\usepackage{bbm}
\usepackage{overpic}
\usepackage{subfig}
\usepackage{wrapfig}
\usepackage[ruled,vlined,linesnumbered]{algorithm2e}
\usepackage{multirow}

\usepackage{color}
\definecolor{green}{rgb}{0, 0.5, 0}
\definecolor{orange}{rgb}{0.8, 0.6, 0.2}
\definecolor{red}{rgb}{1.0, 0.0, 0.0}
\definecolor{teal}{rgb}{0.0, 0.4, 0.4}
\definecolor{purple}{rgb}{0.65,0,0.65}
\definecolor{saffron}{rgb}{0.95,0.75,0.2}
\definecolor{turquoise}{rgb}{0.0,0.5,0.5}
\definecolor{brown}{rgb}{0.5, 0.16, 0.16}

\newcommand{\rz}[1]{{\color{black}#1}}
\newcommand{\rzn}[1]{{\color{black}#1}}

\newcommand{\phil}[1]{{\color{black}{#1}}}
\newcommand{\Lq}[1]{{\color{black}#1}}

\usepackage{overpic}
\usepackage{currfile} 

\newlength\savedwidth

\definecolor{lightgray}{rgb}{0.6, 0.6, 0.6}

\usepackage[normalem]{ulem}

\newcommand{\hidecomment}[1]{}

\usepackage{xspace}

\usepackage[pagebackref=true,breaklinks=true,colorlinks,bookmarks=false]{hyperref}

\threedvfinalcopy 

\def\threedvPaperID{} 

\ifthreedvfinal\fi
\begin{document}

\title{On Learning the Right Attention Point for Feature Enhancement}

\author{
Liqiang Lin$^1$
\quad
Pengdi Huang$^1$
\quad
Chi-Wing Fu$^2$
\quad
Kai Xu$^3$
\quad
Hao Zhang$^4$
\quad
Hui Huang$^1$
\\$^1$Shenzhen University
\\$^2$The Chinese University of Hong Kong
\\$^3$National University of Defense Technology
\\$^4$Simon Fraser University
}

\maketitle

\begin{abstract}
   We present a novel attention-based mechanism to learn enhanced point features for point cloud
processing tasks, e.g.,
classification and segmentation.
\rzn{Unlike prior works\phil{,} which were trained to optimize 
the weights of a pre-selected set of attention points, \phil{our approach learns} to {\em locate\/} the best
%
attention points
%
to maximize the performance of a specific task, e.g., point cloud classification.
%
Importantly, we advocate the use of {\em single\/} 
attention point to facilitate semantic understanding \phil{in point feature learning}.
%
%
%
Specifically, 
\phil{we formulate} 
a new and simple convolution, which combines \rzn{convolutional} features from an input 
%
point and its corresponding {\em learned attention point\/}, or LAP, for short.}
%
Our attention mechanism can be easily incorporated into state-of-the-art point cloud classification and segmentation networks.
Extensive experiments on common benchmarks such as Model-Net40, ShapeNetPart, and S3DIS \phil{all} demonstrate that our \rzn{LAP-enabled} networks consistently outperform the respective original networks, as well as other competitive alternatives, \rzn{which employ multiple attention points, either pre-selected or learned under our LAP framework.}

\end{abstract}

\section{Introduction}
\label{sec:intro}

Learning point features is one of the most fundamental problems in 3D vision and a key building block for
tasks such as shape classification and segmentation.
Conventional convolution employs fixed kernels of varying sizes to aggregate point features, while extensions to variable neighborhoods, which account for anisotropy~\cite{weickert1998} and other local shape properties, have also been studied. 
Another line of approaches follows the non-local means idea~\cite{buades2005non} by collecting features at points that are similar to one another.

\begin{figure}[t!]
    \centering
    \subfloat[Learned attention points (red) for point cloud classification.]{
        \includegraphics[width=\linewidth]{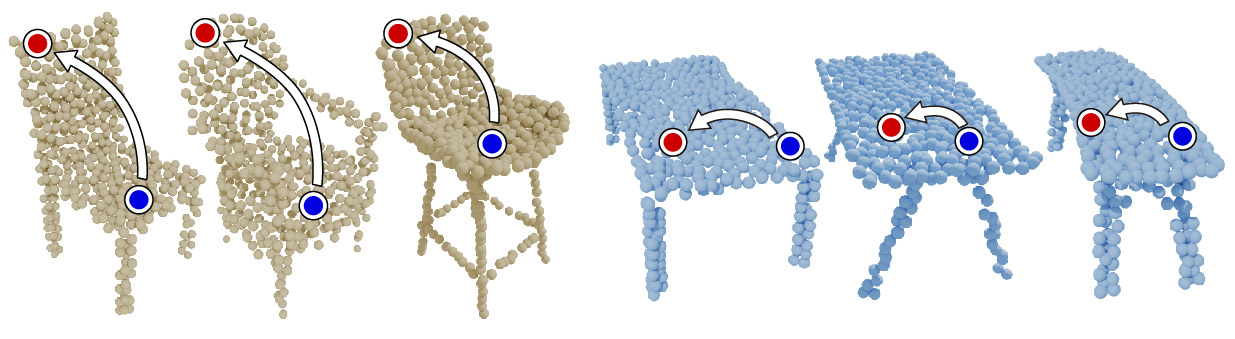}
        \label{fig:teaser:a}
    }\\
    \subfloat[Learned attention points (red) for point cloud segmentation.]{
        \includegraphics[width=\linewidth]{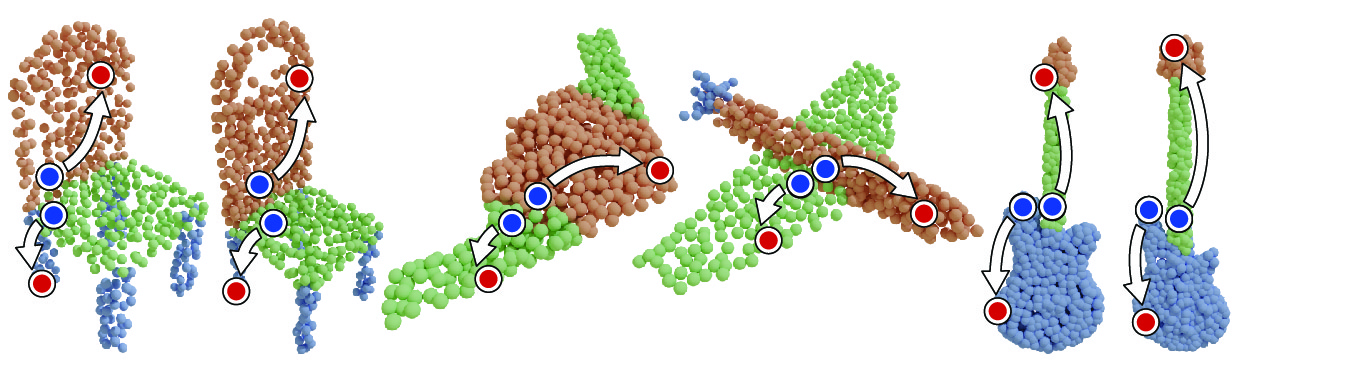}
        \label{fig:teaser:b}
    }\\
    \caption{\rzn{{\em Learned attention points\/}} (LAPs) by our method, shown in red, corresponding to several original (blue) points on two tasks. For classification (a), the LAPs exhibit {\em within-class \rzn{semantic} consistency\/}, while helping to discriminate between different classes: \rzn{attention points for similar original points on chair seats and tabletops are located on different semantic parts.} 
For segmentation (b), \rzn{the two blue points on each shape belong to different semantic regions, but are spatially close\phil{,} thus \phil{sharing} similar spatial neighborhoods. However, their LAPs are located {\em far apart\/} on different \phil{semantic} 
parts, effectively ``pushing'' the blue points into different segments by making their features {\em dissimilar\/}.}}
    \label{fig:teaser}
\end{figure}

Recently, the use of {\em selective attention\/}~\cite{vaswani2017attention,liu2016fully,wang2017residual,peng2018partatt,mnih2014recurrent}
has gained much success in computer vision.
In a typical setting, an attentional network computes the feature of a point by pre-selecting~\cite{hamilton2017inductive,wang2019dynamic,zhao2019pointweb} a set of 
nearest points in the point's neighborhood and learning the associated attention weights to capture additional contextual information through the weights to enrich the point feature.
Another recent approach resorts to finding non-local neighbors~\cite{yan2020pointasnl} by considering points within a much larger neighborhood.

\begin{figure*}[t!]
    \centering
    \includegraphics[width=0.85\linewidth]{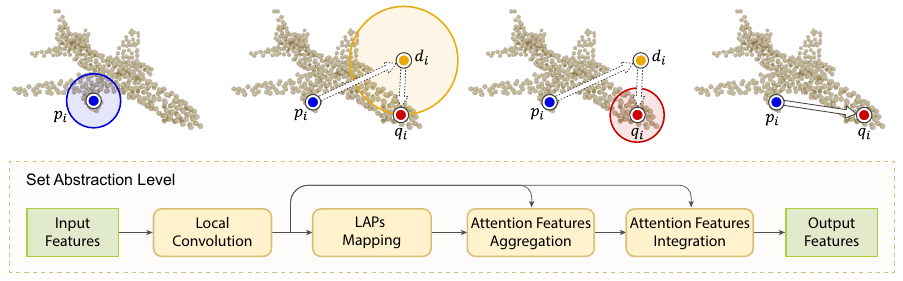}
    \caption{\rzn{Learning to locate the best attention point $q_i$ (red) for a given point $p_i$ (blue). The feature $f_i$ at $p_i$ is updated using arbitrary local convolution. After that, we take $f_i$ to learn an offset vector to obtain $d_i$ (yellow), from which we locate the target attention point $q_i$. We then aggregate the neighboring features of $q_i$, which are finally fused into $p_i$'s features.}
    }
    \label{fig:overview}
\end{figure*}

\begin{figure}[!t]
    \centering
    \subfloat[1 LAP]{
    \includegraphics[width=0.3\linewidth]{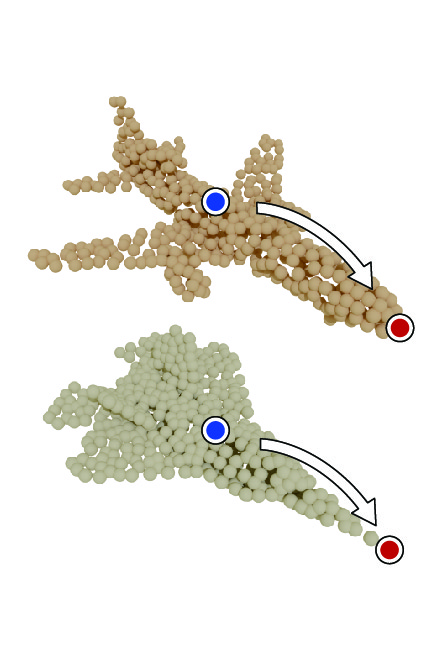}
    }
    \subfloat[2 LAPs]{
    \includegraphics[width=0.3\linewidth]{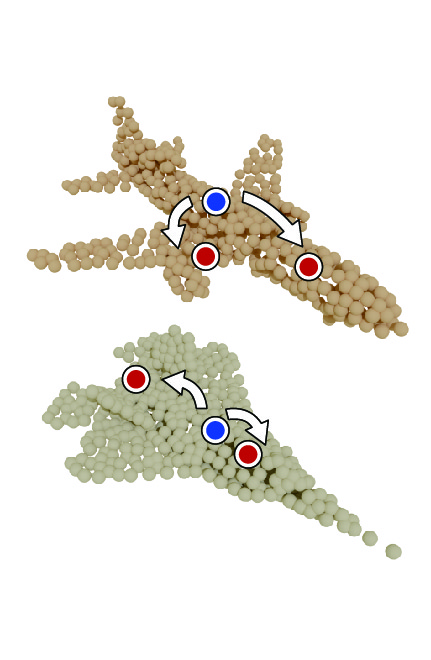}
    }
    \subfloat[4 LAPs]{
    \includegraphics[width=0.3\linewidth]{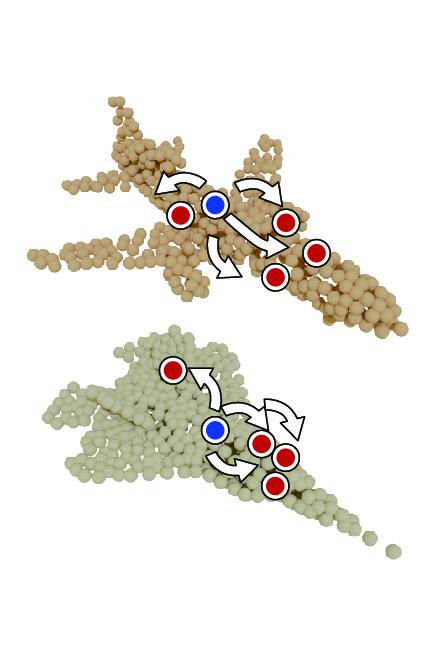}
    }
    \caption{
        Visualizing {\em different number\/} of LAPs (red dots) selected by \rz{the same learning framework, for the same original points (blue dots) on two airplanes.} \rzn{More LAPs, \phil{two or four} 
        in (b) and (c)\phil{,} respectively, tend to exhibit less semantic consistency, compared to the \phil{single} 
        best LAP (a).}
    }
    \label{fig:dap_count}
\end{figure}

In this paper, we present a new model for attentional
point feature learning, which learns to {\em locate a single}
attention point for feature enhancement 
to optimize the performance of a specific point cloud processing task such as
classification and segmentation.
For example, to incorporate our model into a classification network, e.g.,DGCNN~\cite{wang2019dynamic}, that is based on point-wise convolutional feature processing and aggregation, we can readily replace each convolution layer in the network (i.e., EdgeConv in the case of DGCNN) with a new convolution, 
which {\em combines\/} or {\em fuses\/} the convolutional feature at an input point with the convolutional feature at its 
corresponding {\em learned attention point\/} (LAP). Hence, while only one attention point is selected by our method, the feature
fusion captures theinformation from two local point {\em neighborhoods.\/} The LAP is obtained by adding to the original point an offset vector, as 
illustrated in Figure~\ref{fig:overview}. These offset vectors are learned by minimizing the classification errors over the training 
point clouds. 

Unlike prior attentional models, which were trained to optimize the weights of a pre-selected {\em set\/} of attention points~\cite{yan2020pointasnl,zhao2019pointweb} or a {\em sequence\/} via a recurrent network~\cite{liu2019point2sequence}, we focus on learning to {\em locate\/} one best LAP. A key premise 
of our work is that if the right point is found, then adding more attention may not be more effective.
Our intuition is that as shape geometries and structures vary, e.g., within the same class of shapes for classification 
or within the same semantic part in the context of segmentation, it is easier to attain {\em semantic consistency\/} with one LAP than with more, as illustrated in Figure~\ref{fig:dap_count} for a comparison between
one, two, and four LAPs.
Such a consistency implies that matching points on two similar shapes (e.g., the two blue points on the two airplanes in each column of Figure~\ref{fig:dap_count}) tend to be mapped to matching LAPs (the corresponding red points).

As shown in Figure~\ref{fig:teaser}, \rzn{the attention points learned by our method do appear} to exhibit a certain level 
of semantic understanding.
\rzn{Since our training is {\em task-dependent\/},} the LAPs tend to behave differently for different tasks. In particular,
they \phil{may not} possess similar features as the original points. 
A critical observation here is that for 
feature enhancement, points with dissimilar features can be equally useful since
they can provide {\em complementary\/} information.

Our LAP-based attentional point feature learning mechanism can be easily incorporated into modern classification and segmentation networks, 
such as PointNet++~\cite{qi2017pointnet++}, RSCNN~\cite{liu2019relation}, DGCNN~\cite{wang2019dynamic}, and KPConv~\cite{thomas2019kpconv}.
We show through comprehensive tests that, on common benchmarks ModelNet40, ShapeNetPart, 
and S3DIS, our LAP-enabled networks 
{\em consistently\/} improve the performance over the respective original networks, as well as other competitive alternatives, which employ multiple attention points, 
either pre-selected, such as PointASNL~\cite{yan2020pointasnl} and PointWeb~\cite{zhao2019pointweb}, or learned 
under our LAP framework.
The consistent improvements are verified over varying point neighborhood sizes, train-test splits, and ways for
feature integration. 

\section{Related Work}
\label{sec:related}

\paragraph{Deep learning on 3D point clouds.}
Inspired by the seminal work of PointNet~\cite{qi2017pointnet}, many deep neural networks have been developed to directly operate on points~\cite{riegler2017octnet}.
To overcome the inadequacy of PointNet on capturing local structures, PointNet++~\cite{qi2017pointnet++} adopts a deep hierarchical feature learning mechanism that recursively employs PointNet to process point neighborhoods of increasing sizes.
Later, PCNN~\cite{atzmon2018point} adapts
image-based CNNs to the point cloud setting, leading to a permutation-invariant point feature learning.
PointCNN~\cite{li2018pointcnn} 
learns a transformation matrix to weight the input features associated with the input points and rearrange the points into a canonical permutation.
A recent survey on this topic is available in~\cite{guo2020deep}.

One recent work of relevance is S-NET~\cite{dovrat2019sample}, a learning-based point cloud downsampling network that is
also optimized for a particular task (e.g., classification or retrieval), like our work. Also, both their sampling 
problem and our proposed approach to feature enhancement would come down to learning to locate a small number of
points that contribute most to a task at hand. While the success of their work can be seen as validation to our task-specific
training, the problem setup and model design for attention point selection are completely different from S-NET.

\vspace*{-2mm}
\paragraph{Point feature learning.}
Point features play an important role to 
task performance.
So, a majority of deep models on 3D point clouds focus on designing methods to extract better features, typically by exploiting a point's local neighborhood.
DGCNN~\cite{wang2019dynamic} bases the neighboring relations of points on their distances in the feature space and aggregates pair-wise features to generate features for the center point.
PointWeb~\cite{zhao2019pointweb} densely connects every pair of points in a local neighborhood, aiming at extracting point feature that better represents the local region around the center point.

Other methods focus on designing efficient convolution kernels~\cite{boulch2020convpoint, komarichev2019cnn} on points.
PointConv~\cite{wu2019pointconv} treats the convolution kernel as a Monte Carlo estimate of nonlinear functions of the local 3D coordinates of points. Point features are weighted by the estimated density.
KPConv~\cite{thomas2019kpconv} uses a set of points in Euclidean space as the convolution kernel and aggregates input features based on the distances between kernel points and input points.
Liu et al.~\cite{liu2020closerlook3d} reveal that different local operators contribute similarly to 
network performance.
While prior works focus mostly on learning local features, some recent ones~\cite{cheng2020cascaded,yan2020pointasnl} start to explore 
non-local 3D features with the attention mechanism~\cite{vaswani2017attention}.

\vspace*{-2mm}
\paragraph{Attentional point feature learning.}
Attention-based feature learning~\cite{vaswani2017attention} has been introduced to learning point features soon after its applications to image features.
Rather than being exhaustive, we discuss methods that focus on point feature learning rather than on specific vision tasks.
Xie et al.~\cite{xie2018attentional} adopt self-attention to integrate point selection and feature aggregation into a single soft alignment operation.
Yang et al.~\cite{yang2019modeling} propose the point attention transformer, which leverages a parameter-efficient group shuffle attention to learn the point relations.
Zhang et al.~\cite{zhang2019pcan} propose Point Contextual Attention Network to predict the significance of each local point feature based on the point context. 
Chen et al.~\cite{chen2019gapnet} learn local geometric representations by embedding a graph attention mechanism.

Further, attention can be used to learn long-range global features.
Liu et al.~\cite{liu2019point2sequence} propose Point2Sequence, an RNN-based model that captures correlations between different areas in a point cloud.
Lu et al.~\cite{lu2019scanet} design the spatial-channel attention module to capture multi-scale and global context features.
Han et al.~\cite{han2020point2node} create a global graph 
and weights features of distant points with attention.
More recently, Cheng et al.~\cite{cheng2020cascaded} propose global-level blocks to update the feature of a superpoint using weighted features of other superpoints, while
Yan et al.~\cite{yan2020pointasnl} propose PointASNL that samples points over the whole point cloud to query similar ones for non-local point feature learning.

All the methods mentioned above {\em pre-select a set of points\/}, usually points in the local neighborhood or points with similar features, as the attentional points.
In the 2D image domain, Zhang et al.~\cite{zhang2020dynamic} propose to learn an affinity matrix for each attention point to shift it to a better location.
The features of the attention points are weighted based on feature similarity.
On the other hand, Xue et al.~\cite{xue2020not} suggest
that not all attention is needed, and only a small part of the inputs is related to the output targets, through their investigations into several natural language processing tasks.
Our work is inspired by these works which employ learned attentional processing.
Yet, we go beyond them and propose to learn to locate the point to attend to, over the entire point cloud, without relying on feature similarity.
\section{Method}

Recent deep learning approaches for 3D point clouds often focus on designing operators for better local feature extraction. 
Denoting $f_i \in {\mathbb{R}}^{C_1}$ as the input feature of point $p_i$ in a certain network layer ($C_1$ is the channel number), 
this local operator takes the input features of $p_i$'s neighbors and aggregates them to form the output feature $ f'_i \in {\mathbb{R}}^{C_2} $ of $p_i$,
\begin{equation}
\label{eq:basic}
    f'_i = LocalConv(\mathcal{N}(p_i)),
\end{equation}
where $\mathcal{N}(p_i)$ is the set of neighboring points of $p_i$, which are often found by a ball query or k-nearest neighbor (KNN) search.
Here, $LocalConv$ denotes a local convolution.
It can be a point-wise local operator~\cite{qi2017pointnet,qi2017pointnet++}, a grid kernel local operator~\cite{thomas2019kpconv}, or an attention-based local operator~\cite{zhao2019pointweb}.

Some methods use attention to learn global features for long-range structures, in which $\mathcal{N}(p_i)$ is replaced by a set of pre-selected attention points.
These points can be found by locating points with similar features as $f_i$ in the feature space.
Like Eq.~\eqref{eq:basic}, a basic mechanism is to weigh the input features of the attention points based on the feature similarity, then to combine them into the output feature of $p_i$.

In this work, we propose a new network module, called LAP-Conv, that {\em directly learns to find\/} attention points without relying on the feature similarity.
In particular, we learn to find for each input point $p_i$ {\em only one\/} attention point instead of finding multiple ones.
The top part of Figure~\ref{fig:overview} illustrates 
how LAP-Conv works.
Let $P$ be the input cloud with $n$ points $\{ p_1, p_2, ..., p_n\}$.
Taking as an example the blue point in Figure~\ref{fig:overview} as $p_i$,
we learn to find its associated attention point $q_i$, the red point, through the assistance of the offset point $d_i$, the yellow point.
The bottom part of Figure~\ref{fig:overview} shows the overall feature learning pipeline.
First, we update feature $f_i$ of point $p_i$ using a Local Convolution block (Eq.~\eqref{eq:basic}), then map $f_i$ to the corresponding attention point $q_i$ using the LAPs Mapping block (Section~\ref{sec:method:learn_daps}).
Next, the Attention Features Aggregation block (Section~\ref{sec:method:agg_feat}) aggregates features for the attention point $q_i$.
Finally, the Attention Features Integration block (Section~\ref{sec:method:int_feat}) integrates the feature of the attention point into the output feature of $p_i$. Also, we show the detailed structure of LAP-Conv in Figure~\ref{fig:arch}, in which the three blocks are outlined.

\subsection{Learning to Locate Attention Points}
\label{sec:method:learn_daps}

The features of points in the point cloud are first updated with Eq.~\eqref{eq:basic}.
We learn a function $D$ to map the input feature of point $p_i$ to locate offset point $d_i$ (the yellow point in Figure~\ref{fig:overview}) with
    $d_i = D(f_i)$, 
and then find the target attention point $q_i$ (the red point) with the assistance of point $d_i$.
Inspired by the way how neighboring points are found in the feature space~\cite{wang2019dynamic}, we propose to learn to find the attention points either in the Euclidean space or in the feature space.

{\em Case (i)\/}.
For attention points in the Euclidean space, we use a small MLP to map the feature $f_i$ to a three-dimensional offset vector in the Euclidean space. We then add this vector back to the 3D coordinates $x_i$ of the original point $p_i$ to locate the offset point
\begin{equation}
    d_i = MLP(f_i) + x_i.
    \label{eq:dap1}
\end{equation}
This MLP is shared among all the points in the input point cloud $P$. Also, point $d_i$ is not necessarily a point in $P$. Its location is arbitrary in the Euclidean space. Hence, we map the feature of point $p_i$ to an offset vector, which acts as an attention direction for further pinpointing $d_i$. Taking point $d_i$ as 
guidance, we then find the nearest point $q_i$ to point $d_i$ among all the points in the entire point cloud as the learned attention point $q_i$ associated with $p_i$.

\begin{figure}[t]
    \centering
    \includegraphics[width=0.99\linewidth]{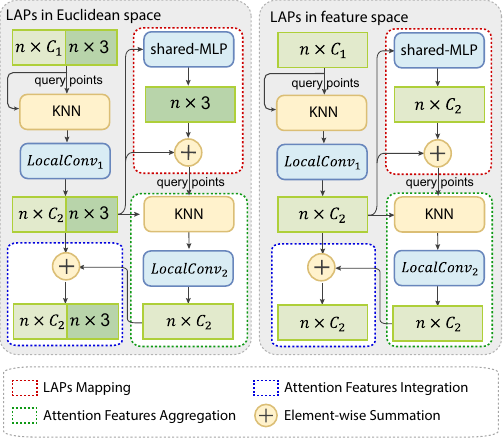}
    \caption{The structure of LAP-Conv. For each attention point in the Euclidean space, we learn a 3-dimensional offset vector and add it back. Each resulting point is used as a query point to find the k nearest points in the point cloud. The features of these points are aggregated into the learned attention point and the feature of each attention point is integrated into the original feature. For attention points in the feature space, the offset vectors are in the same space and the neighboring points are also found in the feature space.}
    \label{fig:arch}
\end{figure}

{\em Case (ii)\/}.
For attention points in the feature space, the feature $f_i$ of point $p_i$ is mapped via a small shared MLP to an offset vector (with same channel number as $f_i$) in the feature space.
This vector is then added to $f_i$ to produce $f_{d_i}$, which is an offset feature in the feature space:
\begin{equation}
    f_{d_i} = MLP(f_i) + f_i.
    \label{eq:dap2}
\end{equation}
Similar to $d_i$ in case (i), offset feature $f_{d_i}$ is not necessarily a feature vector of the points in the original point cloud. Also, it can have arbitrary values.
With $f_{d_i}$, we then search the entire point cloud for point $q_i$ with the nearest feature vector $f_{q_i}$ to $f_{d_i}$ in the feature space, and take $q_i$ as the corresponding learned attention point\/ for point $p_i$.
The blocks with the red-dotted frames in Figure~\ref{fig:arch} illustrate the detailed procedure of this LAPs Mapping.

\subsection{Attentional Features Aggregation}
\label{sec:method:agg_feat}
After finding the target attention point $q_i$ for point $p_i$, we then aggregate the features of the neighborhood of $q_i$ and update the feature of point $q_i$ with
\begin{equation}
    f_{q_i} = LocalConv(\mathcal{N}(q_i)).
    \label{eq:agg1}
\end{equation}
For attention points in the Euclidean space, these neighboring points are found in the Euclidean space. 
For attention points in the feature space, these neighboring points are also found in the feature space. 

To find the closest point $q_i$ to $d_i$, we need to calculate the distances between point $d_i$ and all the other points. To further find the neighboring points of $q_i$, another distance calculation is required.
To reduce the cost of processing two sets of distance calculations, we directly find the neighboring points of point $d_i$ instead of explicitly finding point $q_i$ and then its neighbors.
The learned attention point $q_i$ must be among the neighboring points of $d_i$, since $q_i$ is the nearest one to $d_i$ in the original point cloud.
So, the feature of the learned attention point can be updated using the features of the neighboring points of $d_i$ as
\begin{equation}
    f_{q_i} = LocalConv( \mathcal{N} (d_i) ).
    \label{eq:agg2}
\end{equation}
For both cases (i) and (ii),
we prefer to find these neighboring points using KNN. In this situation, no matter how large the offset vectors are, there are always neighboring points of $d_i$. If the neighboring points are found with a ball query~\cite{qi2017pointnet++}, an additional loss is needed to punish the point $d_i$ from being shifted too far away. Otherwise, there could be no neighboring points of $d_i$ within a certain radius.

Hence, we use KNN to finding the neighboring points for performing $LocalConv_1$ for consistency. 
After we learn the offset points, we use them as query points to find k neighboring points and update the feature of $q_i$, as shown in the green dotted frames in Figure~\ref{fig:arch}.
It should be noticed that all these neighboring points are points from the original point cloud (case (i)) or the features of the points in the original point cloud (case (ii)), in which $d_i$ or $f_{d_i}$ is not one of them.

\subsection{Integration of Attentional Features}
\label{sec:method:int_feat}

After we aggregate a feature for the learned attention point $q_i$, we then integrate the feature into the feature of $p_i$ with an integration function $I$ (`I' for integration):
\begin{equation}
    f'_i = I( LocalConv_1(\mathcal{N}(p_i)), LocalConv_2(\mathcal{N}(d_i))),
\end{equation}
where we could use the same or different convolutional operators for $LocalConv_1$ and $LocalConv_2$.

In the previous attention-based networks for processing point clouds,~\eg,~\cite{wang2019dynamic,zhao2019pointweb,cheng2020cascaded,yan2020pointasnl,yang2019modeling,han2020point2node,xie2018attentional,liang20193d},
the features of the attention points are weighted by a dot-product similarity between the features of the pre-selected attention points and $p_i$, or by the similarity in the spatial location between them.
An attention point with a more similar feature is given a larger weight when added into the feature of $p_i$.
However, we argue that
\emph{not only\/} points with similar features could be useful, \emph{but also\/} points with \emph{different features\/} could also be useful in 
feature learning.
Particularly, points with different features could supply the center point $p_i$ with \emph{vital context for performing the target task.\/}

Here, one straightforward choice for $I$ is to simply add the features of point $q_i$ and point $p_i$ together:
\begin{equation}
    I = Add( f_{p_i} , f_{q_i} ),
    \label{eq:i1}
\end{equation}
as illustrated in the blue dotted frames in Figure~\ref{fig:arch}.
Since $LocalConv_1$ and $LocalConv_2$ do not share parameters, the network can learn to update the feature of $q_i$ for adding it into the feature of $p_i$.
Another choice for $I$ is to concatenate the two features then use an MLP to reduce the dimensions:
\begin{equation}
    I = MLP( Concatenate( f_{p_i} , f_{q_i} )).
    \label{eq:i2}
\end{equation}
A comparison result between these two forms of feature integration can be found in Table~\ref{tab:i} of Section~\ref{sec:results}.

\section{Results and Evaluation}
\label{sec:results}

Our proposed LAP-Conv can be easily adapted into conventional point-wise feature learning networks.
In existing networks, a $LocalConv$ block is generally used in a set abstraction level to update the input feature $f_i$ of a point and produce $f'_i$, as described in Eq.~\eqref{eq:basic}.
Hence, by keeping the channel sizes of the input and output features to be the same, we can replace the $LocalConv$ block in existing networks with our LAP-Conv to boost the network performance.

To demonstrate the effectiveness of our LAP-Conv, we adapt it into common networks, including PointNet++~\cite{qi2017pointnet++}, RSCNN~\cite{liu2019relation}, DGCNN~\cite{wang2019dynamic}, and KPConv~\cite{thomas2019kpconv}, producing LAP-enabled networks,~\ie, LAP-PointNet++, LAP-RSCNN, LAP-DGCNN, and LAP-KPConv.
Note that since DGCNN finds neighboring points in feature space, the attention points for LAP-DGCNN are learned in the feature space (Eq.~\eqref{eq:dap1}). For the other LAP-enabled networks, the LAPs are learned in Euclidean space (Eq.~\eqref{eq:dap2}).

We conduct experiments using these networks on various tasks, including shape classification (Section~\ref{sec:results:cls}), part segmentation, and semantic segmentation (Section~\ref{sec:results:seg}), showing both quantitative comparisons and qualitative results.
At last, we evaluate various aspects of our LAP-Conv (Section~\ref{sec:results:abl}),~\eg, one vs.~multiple learned attention point\phil{s}, neighborhood sizes, train/test split, etc.


\subsection{Classification}
\label{sec:results:cls}
\begin{table}
    \centering
    \caption{3D Shape classification results on ModelNet40.
        OA: overall accuracy (\%);
        mAcc: mean class accuracy (\%);
        xyz: points only as input;
        xyz+nor: use points and normals. }
    \label{tab:mn40}
    \begin{tabular}{c|ccc}
        \hline
        Method                              & input   & \#points & OA            \\
        \hline \hline

        PointNet~\cite{qi2017pointnet++}    & xyz     & 1k       & 89.2          \\
        O-CNN~\cite{wang2017cnn}            & xyz+nor & -        & 90.6          \\
        PAT~\cite{yang2019modeling}         & xyz+nor & 1k       & 91.7          \\
        Kd-Net~\cite{klokov2017escape}      & xyz     & 32k      & 91.8          \\ 
        PointCNN~\cite{li2018pointcnn}      & xyz     & 1k       & 92.2          \\
        PCNN~\cite{atzmon2018point}         & xyz     & 1k       & 92.3          \\
        PointWeb~\cite{zhao2019pointweb}    & xyz+nor & 1k       & 92.3          \\
        SpiderCNN~\cite{xu2018spidercnn}    & xyz+nor & 5k       & 92.4          \\
        PointConv~\cite{wu2019pointconv}    & xyz+nor & 1k       & 92.5          \\
        A-CNN~\cite{komarichev2019cnn}      & xyz     & 1k       & 92.6          \\
        Point2Node~\cite{han2020point2node} & xyz     & 1k       & 93.0          \\
        PointASNL~\cite{yan2020pointasnl}   & xyz     & 1k       & 92.9          \\
        PointASNL~\cite{yan2020pointasnl}   & xyz+nor & 1k       & 93.2          \\
        DensePoint~\cite{liu2019densepoint} & xyz     & 1k       & 93.2          \\
        SO-Net~\cite{li2018so}              & xyz+nor & 5k       & 93.4          \\
        \hline
        PointNet++~\cite{qi2017pointnet++}  & xyz     & 1k       & 90.7          \\
        RSCNN~\cite{liu2019relation}        & xyz     & 1k       & 91.7          \\
        KPConv~\cite{thomas2019kpconv}      & xyz     & 7k       & 91.9     \\
        DGCNN~\cite{wang2019dynamic}        & xyz     & 1k       & 92.9          \\
        \hline
        LAP-PointNet++                      & xyz     & 1k       & 92.9          \\
        LAP-RSCNN                           & xyz     & 1k       & 92.8          \\
        LAP-KPConv                          & xyz     & 7k       & 92.3          \\
        \textbf{LAP-DGCNN}                  & xyz     & 1k       & \textbf{93.9} \\
        \hline
    \end{tabular}
\end{table}

First, we evaluate our method on the shape classification task using ModelNet40~\cite{wu20153d}, which provides a comprehensive collection of
12,311 CAD models from 40 categories.
Here, we use the input point clouds sampled from these models by PointNet~\cite{qi2017pointnet}.
Typically, 1,024 points are uniformly sampled per model to serve as the network input.
Following the official train/test split, we use 9,843 models for training and 2,468 models for testing.

In this experiment, we evaluate the performance of LAP-PointNet++, LAP-RSCNN, LAP-KPConv, and LAP-DGCNN, where data preprocessing, augmentation, and the choice of hyper-parameters all follow the original network settings. Also, we employ only 3D point coordinates as network inputs.
For LAP-PointNet++, 
$LocalConv_1$ and $LocalConv_2$ in Figure~\ref{fig:arch} are small shared MLPs with max-pooling separately. 
For LAP-RSCNN, 
$LocalConv_1$ is an RS-Conv and $LocalConv_2$ is a small shared-MLP with max-pooling.
Note that, since only the single-scale version of the RSCNN code is officially released, we develop our LAP-RSCNN based on the released code and report 91.7 for the single-scale RSCNN, following~\cite{li2020pointaugment}.
For LAP-KPConv, 
$LocalConv_1$ is a KP-Conv and $LocalConv_2$ is a small shared-MLP with max-pooling.
In this paper, we report 91.9 for KPConv as the result reproduced by the officially released code.

Finally, for LAP-DGCNN, 
we use EdgeConv for both $LocalConv_1$ and $LocalConv_2$.

Table~\ref{tab:mn40} reports the results of using different methods for the shape classification task. Comparing the results of the three LAP-enabled networks with those of the original counterparts, we can see that all of them improve the performance over the original networks, showing that our LAP-Conv can help improve the quality of the features for the shape classification task.
Overall, LAP-DGCNN attains the highest performance with an overall accuracy of 93.9 to date on the classification task, compared with all others,
including other attention-based methods such as PointWeb~\cite{zhao2019pointweb}, Point2Node~\cite{han2020point2node}, and PointASNL~\cite{yan2020pointasnl}.

\begin{figure}[ht]
    \centering
    \subfloat[Ground-truth]{
        \includegraphics[width=0.3\linewidth]{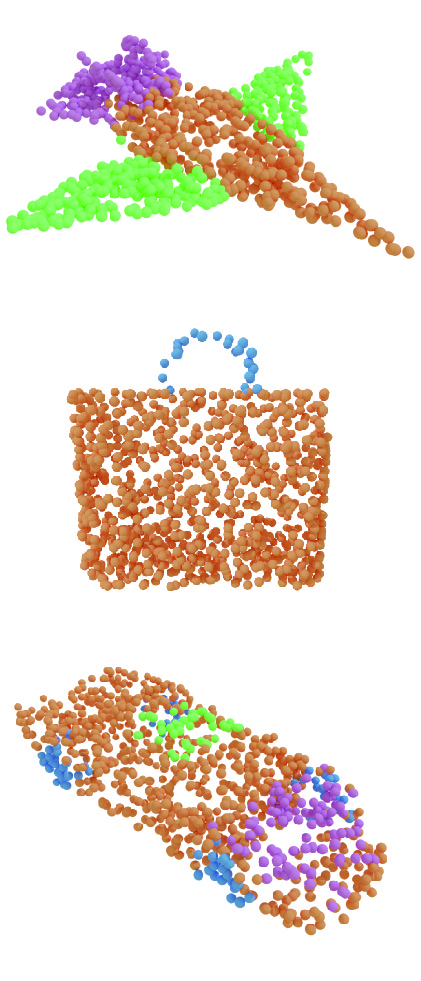}
    }
    \subfloat[PointNet++]{
        \includegraphics[width=0.3\linewidth]{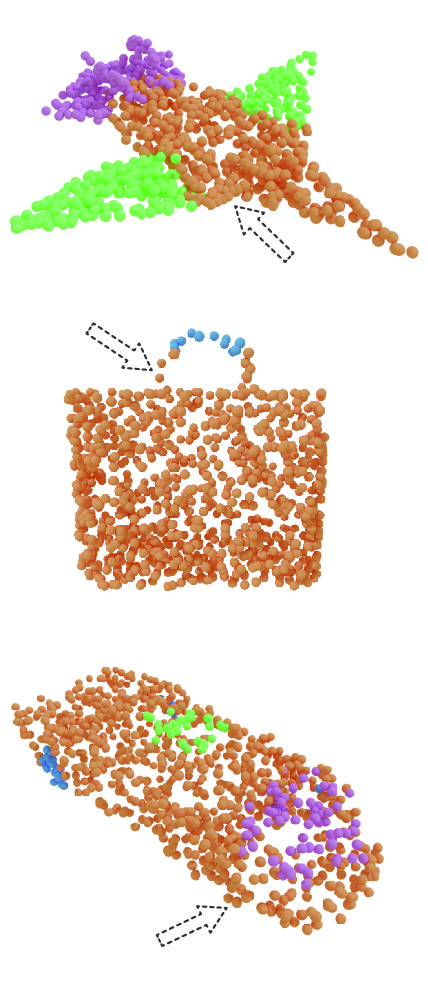}
    }
    \subfloat[LAP-PointNet++]{
        \includegraphics[width=0.3\linewidth]{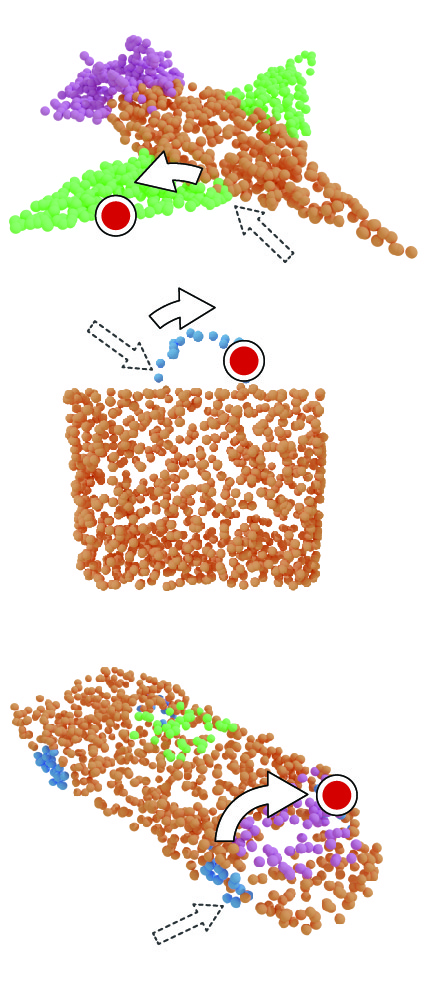}
    }
    \caption{Comparing part segmentation on ShapeNet Part between PointNet++ and LAP-PointNet++.
        In each case, we highlight an incorrectly labeled point near a segmentation boundary ({\em hollow\/} arrow) by PointNet++ and how the use of LAP-Conv (right) corrected the label with feature enhancement by an attention point (large red dot).}
    \label{fig:partseg}
\end{figure}

\subsection{Segmentation}
\label{sec:results:seg}

\begin{table}
    \centering
    \caption{Part segmentation results on ShapeNet Part.}
    \label{tab:partseg}
    \begin{tabular}{cc}
        \hline
        Method                             & mIoU          \\
        \hline \hline
        Kd-Net~\cite{klokov2017escape}     & 82.3          \\
        PointNet~\cite{qi2017pointnet}     & 83.7          \\
        PointNet++~\cite{qi2017pointnet++} & 85.1          \\
        PCNN~\cite{atzmon2018point}        & 85.1          \\
        DGCNN~\cite{wang2019dynamic}       & 85.2          \\
        RSCNN~\cite{liu2019relation}       & 86.2          \\ 
        KPConv~\cite{thomas2019kpconv}     & 86.4          \\
        \hline
        LAP-PointNet++                     & 85.4          \\
        LAP-DGCNN                          & 85.8          \\
        LAP-RSCNN                          & 86.4          \\
        \textbf{LAP-KPConv}                & \textbf{86.9} \\
        \hline
    \end{tabular}
\end{table}

\paragraph{Part segmentation.}
The ShapeNet Part dataset~\cite{yi2016scalable}
has 16,881 shapes over 16 categories and with 50 part labels. 
We sample 2,048 points per shape from this dataset for our experiments and follow the official data split setup~\cite{chang2015shapenet} adopted in DGCNN~\cite{wang2019dynamic} and PointNet++~\cite{qi2017pointnet++}.

Table~\ref{tab:partseg} reports the performance by various method\phil{s}, in terms of the mean Intersection-over-Union (mIoU) metric, showing that LAP-Conv helps improve the part segmentation performance of PointNet++, RSCNN, DGCNN, and KPConv. Figure~\ref{fig:partseg} visually contrasts PointNet++ and LAP-PointNet++ to show the differences
made by LAP-Conv; see more such results in the supplementary material.

\begin{table}[!t]
    \centering
    \caption{Semantic segmentation results on the S3DIS dataset (evaluated on Area 5).}
    \label{tab:semseg}
    \begin{tabular}{cc}
        \hline
        Method                                    & mIoU          \\
        \hline \hline
        PointNet~\cite{qi2017pointnet}            & 41.1          \\
        DGCNN~\cite{wang2019dynamic}              & 49.1          \\
        TangentConv~\cite{tatarchenko2018tangent} & 52.6          \\
        PointCNN~\cite{li2018pointcnn}            & 57.3          \\
        SPGraph~\cite{landrieu2018large}          & 58.0          \\
        ParamConv~\cite{wang2018deep}             & 58.3          \\
        PointWeb~\cite{zhao2019pointweb}          & 60.3          \\
        HPEIN~\cite{jiang2019hierarchical}        & 61.9          \\
        MVPNet~\cite{jaritz2019multi}             & 62.4          \\
        Point2Node~\cite{han2020point2node}       & 63.0          \\
        MinkowskiNet~\cite{choy20194d}            & 65.4          \\
        KPConv~\cite{thomas2019kpconv}            & 67.1          \\
        JSENet~\cite{hu2020jsenet}                & 67.7          \\
        \hline
        LAP-DGCNN                                 & 53.6          \\
        \textbf{LAP-KPConv}                       & \textbf{68.2} \\
        \hline
    \end{tabular}
\end{table}

\vspace*{-3.5mm}
\paragraph{Indoor scene semantic segmentation.}
We use the Stanford 3D Large-Scale Indoor Spaces (S3DIS) dataset~\cite{armeni20163d}, which contains large-scale 3D-scanned point clouds for six indoor areas with 272 rooms from three different buildings.
Altogether, the dataset has $\sim$ 273 million points, each belonging to one of 13 semantic categories. Area-5 is used as the test scene for evaluating the method's generalizability.

\begin{figure}[!t]
    \centering
    \includegraphics[width=0.99\linewidth]{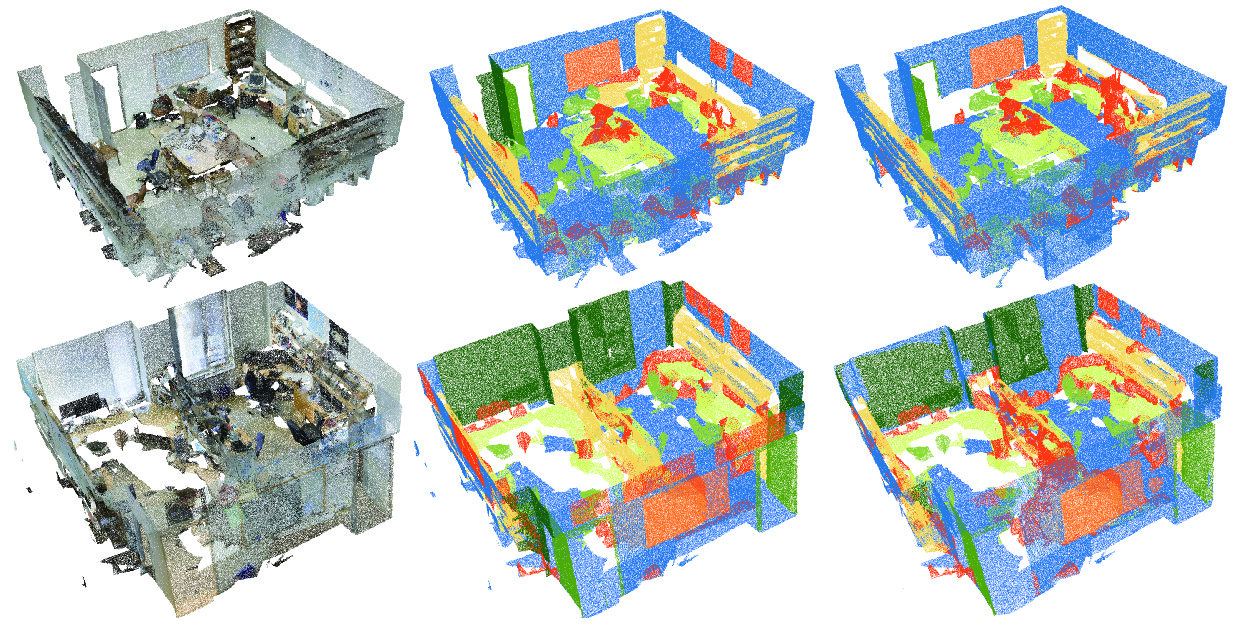}
    \vspace*{-1.5mm}
    \caption{Visualization of two semantic segmentation results on S3DIS.
        Left: input point cloud with RGB colors.
        Middle: points with ground-truth semantic labels.
        Right: segmentation results predicted by LAP-DGCNN.}
    \label{fig:semseg}
    \vspace*{-2mm}
\end{figure}

For LAP-DGCNN, each room is split into blocks of size $1m$$\times$$1m$ as done in DGCNN. The input is point coordinates together with RGB colors and normalized spatial coordinates (a 9D vector per point), following the settings in DGCNN~\cite{wang2019dynamic}.
During the training, we sample 4,096 points from each room block.
Then, all sampled points are used for testing.
%
For LAP-KPConv, 
each 3D scene in the dataset is segmented into small sub-clouds contained in spheres.
During the training, the spheres were picked randomly in the scenes, and during testing, they are picked regularly in the point clouds.
Table~\ref{tab:semseg} gives the results, showing that LAP-Conv helps improve semantic segmentation performance on S3DIS for both DGCNN and KPConv. 
Also, we present two visual semantic segmentation results on S3DIS in Figure~\ref{fig:semseg}.

Note that since PointNet++ and RSCNN did not report their results on S3DIS, LAP-PointNet++ and LAP-RSCNN are not evaluated on S3DIS.

\subsection{Ablation Study}
\label{sec:results:abl}
Finally, we evaluate various aspects of our method, employing
LAP-DGCNN to test on ModelNet40.

\vspace*{-3.5mm}
\paragraph{Different ways of choosing LAPs.}
In this experiment, we compare LAP-DGCNN with three variants of the network in terms of how to choose attention points:
\begin{itemize}
    \item Random-LAP-DGCNN, in which we replace the MLP for learning to generate the offsets (Eq.~\eqref{eq:dap2}) 
          with a randomly-generated vector, thus replacing the core of our LAP-Conv, which uses the learned offset vector to locate the learned attention point;
    \item LAP2-DGCNN and LAP4-DGCNN, which respectively learns two and four attention points simultaneously, instead of one, in one single LAP-Conv module.
\end{itemize}

As comparison results in Table~\ref{tab:abla} show, random-LAP-DGCNN hardly makes any improvement over the baseline, which is DGCNN.
This result demonstrates the importance of the learned offset vectors for searching for the target attention point.
Yet, using just the right learned attention point, as in LAP-DGCNN, we can achieve a large performance improvement over the original DGCNN.

Next, comparing LAP2-DGCNN and LAP4-DGCNN with LAP-DGCNN, we can see that learning to find and use more than one attention point improves over DGCNN.
However, the improvement is far below than using just one single learned attention point.
Yet, an interesting observation is that the more learned attention points we use, resulting in larger network capacity, the worse the results appear to be,
\Lq{indicating that LAP's effectiveness is not merely brought by more convolution layers.}
The visualization in Figure~\ref{fig:dap_count} on where the 1, 2, and 4 attention points were selected by the networks may
hint at a possible reason: as more points are added, their consistency tends to drop.

\begin{table}
    \centering
    \caption{Comparing different ways of choosing LAPs.}
    \label{tab:abla}
    \begin{tabular}{@{\hspace*{1mm}}c@{\hspace*{3mm}}c@{\hspace*{3mm}}c@{\hspace*{3mm}}c@{\hspace*{1mm}}}
        \hline
                           & OA            & mAcc          & \Lq{Parameters (M)} \\
        \hline \hline
        DGCNN (baseline)   & 92.9          & 90.2          & \Lq{4.21}           \\
        random-LAP-DGCNN   & 93.0          & 89.9          & \Lq{4.25}           \\
        \textbf{LAP-DGCNN} & \textbf{93.9} & \textbf{91.5} & \Lq{4.28}           \\
        LAP2-DGCNN         & 93.3          & 90.2          & \Lq{4.35}           \\
        LAP4-DGCNN         & 93.1          & 90.3          & \Lq{4.49}           \\
        \hline
    \end{tabular}
\end{table}

\vspace*{-3.5mm}
\paragraph{Different feature integration.}
We can have different ways of integrating the feature of point $p_i$ with that of its LAP $q_i$; see Section~\ref{sec:method:int_feat}.
As shown in Table~\ref{tab:i}, option $I_1$ (Eq.~\eqref{eq:i1}) gives a slightly better performance compared with $I_2$ (Eq.~\eqref{eq:i2}), so we empirically choose $I$ in our LAP-Conv.
We have tested other ways of feature integration,~\eg, point-wise multiply, but they do not lead to good results.

\begin{table}
    \centering
    \caption{Comparing different ways of integrating features.}
    \label{tab:i}
    \begin{tabular}{ccc}
        \hline
        Different Function $I$     & OA   & mAcc \\
        \hline \hline
        $I_1$ or Eq.~\eqref{eq:i1} & 93.9 & 91.5 \\
        $I_2$ or Eq.~\eqref{eq:i2} & 93.6 & 91.0 \\
        \hline
    \end{tabular}
\end{table}

\begin{table}
    \centering
    \caption{Comparing different neighboring sizes ($k$) in KNN.}
    \label{tab:k}
    \begin{tabular}{ccc}
        \hline
        \multirow{2}{*}{$k$} & \multicolumn{2}{c}{OA}                 \\
        \cline{2-3}          & DGCNN                  & LAP-DGCNN     \\
        \hline \hline
        5                    & 90.5                   & \textbf{93.5} \\
        10                   & 91.4                   & \textbf{93.4} \\
        20                   & 92.9                   & \textbf{93.9} \\
        40                   & 92.4                   & \textbf{93.3} \\
        \hline
    \end{tabular}
\end{table}

\begin{table}
    \centering
    \caption{Shape classification results on ModelNet40 with different train/test splits.}
    \label{tab:splits}
    \begin{tabular}{ccc}
        \hline
        \multirow{2}{*}{(Train/all)\%} & \multicolumn{2}{c}{OA}                 \\
        \cline{2-3}                    & DGCNN                  & LAP-DGCNN     \\
        \hline \hline
        80\%                           & 92.9                   & \textbf{93.9} \\
        60\%                           & 91.5                   & \textbf{92.6} \\
        40\%                           & 90.6                   & \textbf{92.1} \\
        20\%                           & 88.9                   & \textbf{90.2} \\
        10\%                           & 87.3                   & \textbf{88.0} \\
        5\%                            & 82.7                   & \textbf{83.1} \\
        1\%                            & 63.5                   & \textbf{64.9} \\
        \hline
    \end{tabular}
\end{table}

\vspace*{-3.5mm}
\paragraph{Different neighborhood sizes.}
Note that we built LAP-DGCNN by following the setting of DGCNN to use $k$$=$$20$ for the two KNN operations in LAP-Conv. In this experiment, we test LAP-DGCNN with different neighboring sizes ($k$), while keeping all the other settings unchanged.

As shown by results in Table~\ref{tab:k}, the performance of our LAP-DGCNN is quite robust against different choices of $k$, and it also consistently outperforms DGCNN. Even with a rather small neighborhood size $k$$=$$5$, our method can already attain a rather high performance, which exceeds that of DGCNN for all the neighborhood sizes shown in Table~\ref{tab:k}, {\em as well as} all other methods listed in Table~\ref{tab:mn40}.

\vspace*{-3.5mm}
\paragraph{Different train/test splits.}
In a final experiment, we test the robustness of our method over different training/test splits. Note that the original setting uses 9,843 models, which is around 80\% of the whole dataset,
for training, and the remaining 20\% for testing.
We test the performance of our LAP-DGCNN using smaller training sets, from 80\% down to 1\% of the whole dataset, and compare to DGCNN, with results shown in Table~\ref{tab:splits}. Similar to~\cite{sauder2019self}, we select the training samples as follows, for all train/test splits:
we randomly sample one object per class first; the remaining training data are randomly sampled from the original training set, regardless of their classes.
As we can observe from Table~\ref{tab:splits}, our method performs well even with a very small training set, and most importantly, it
consistently outperforms DGCNN over all train/test splits.
Also, the randomization during our setup of the multiple training sets can alleviate potential concerns of over-fitting.

\section{Discussion, Limitation, and Future Work}
\label{sec:future}

\begin{figure}[!t]
   \centering
   \subfloat[]{
      \includegraphics[width=0.26\linewidth]{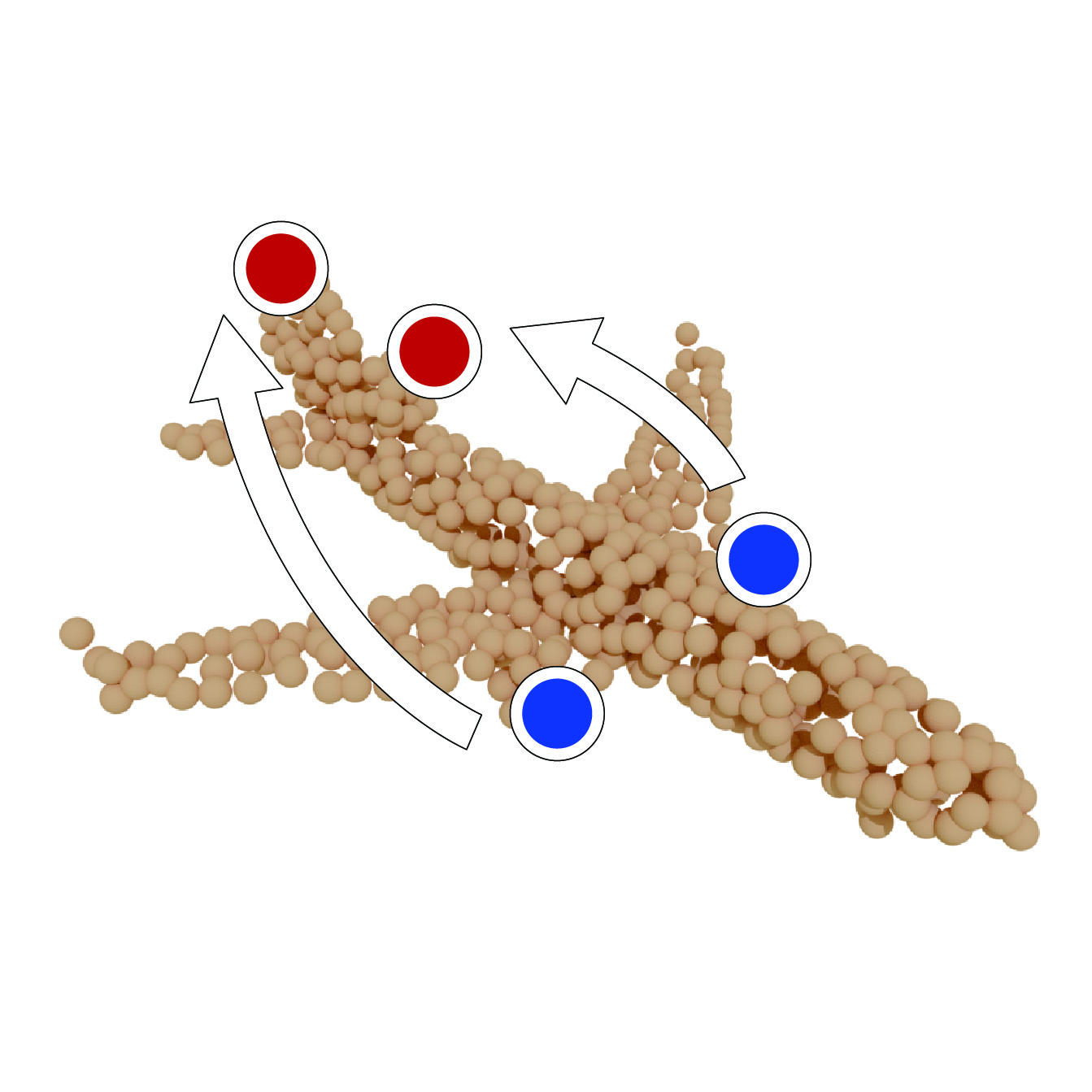}
   }
   \hspace*{-3mm}
   \subfloat[]{
      \includegraphics[width=0.23\linewidth]{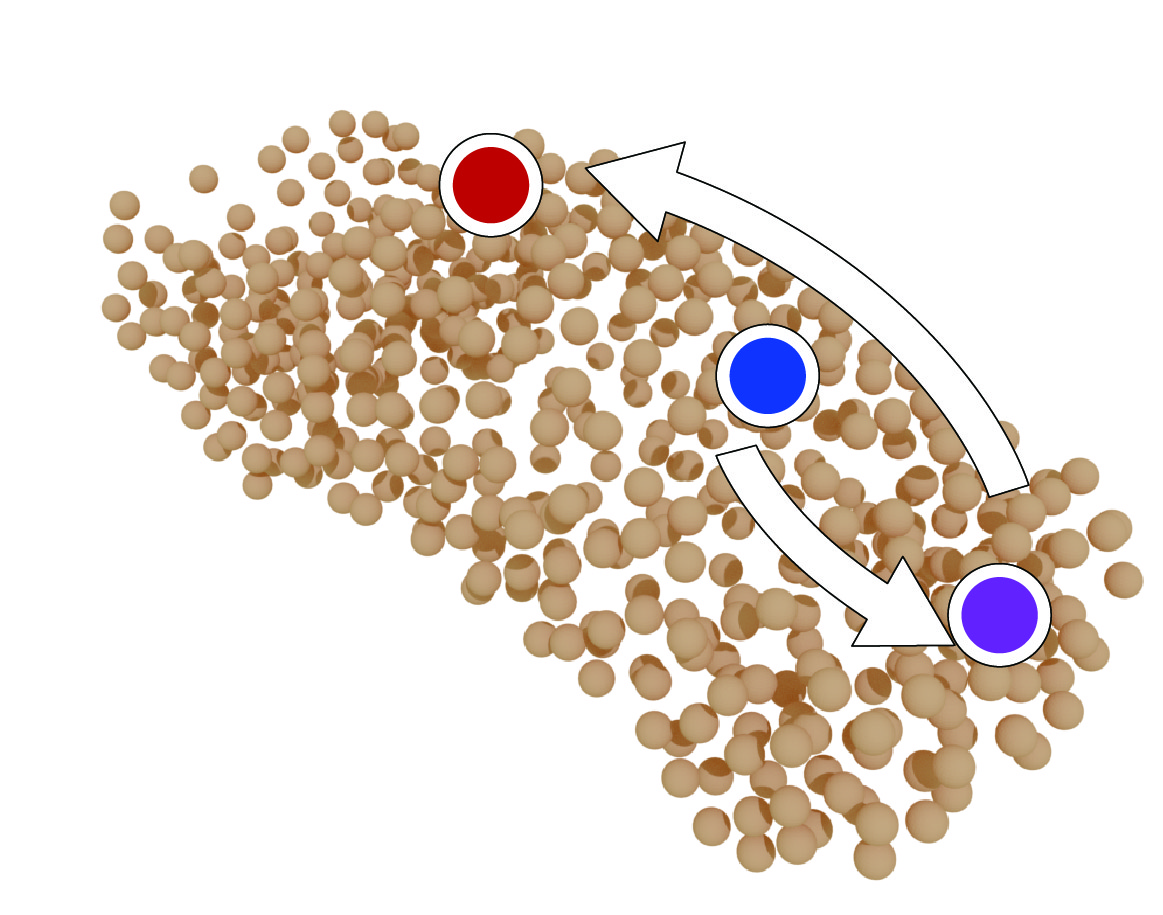}
   }
   \hspace*{-3mm}
   \subfloat[]{
      \includegraphics[width=0.26\linewidth]{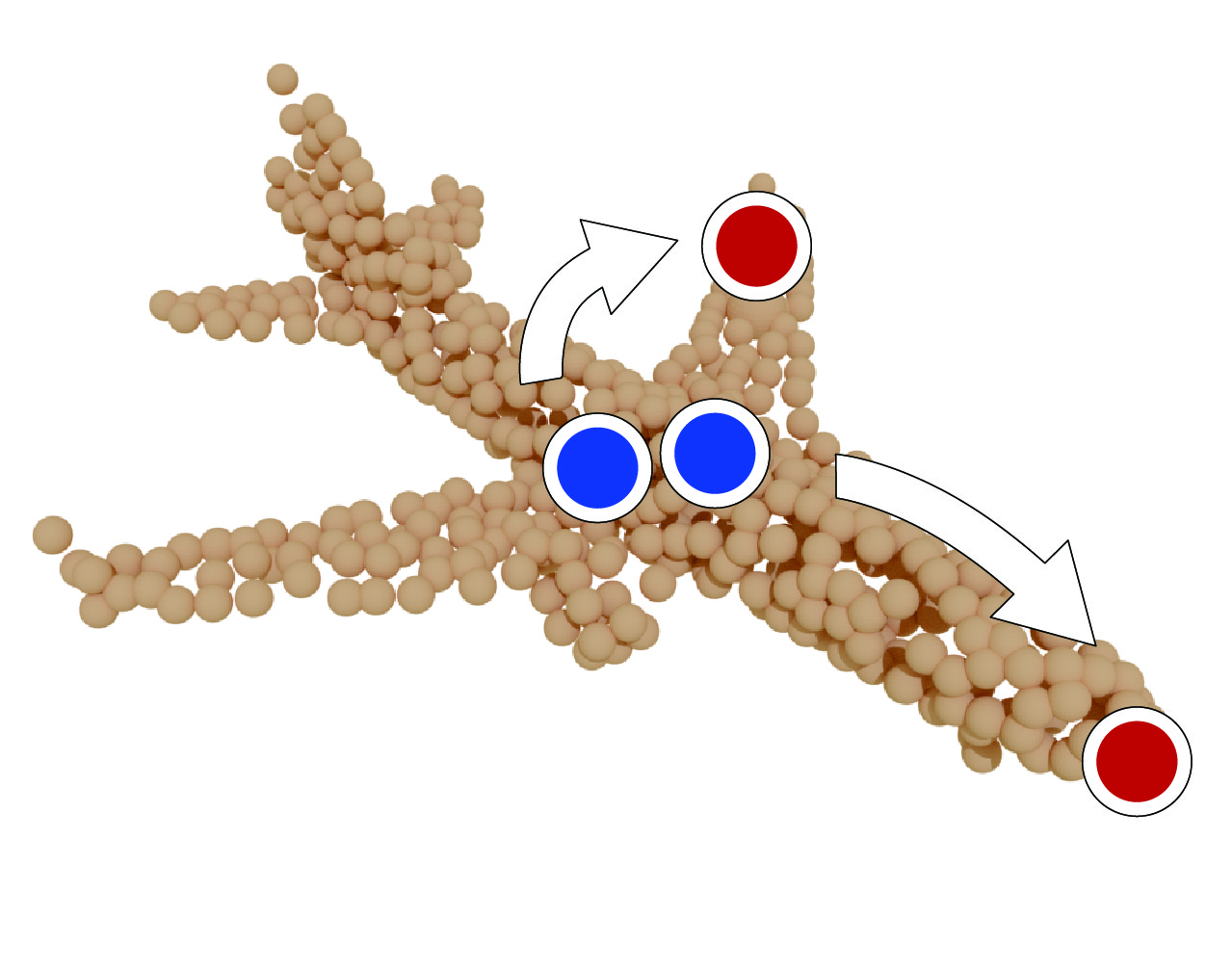}
   }
   \hspace*{-3mm}
   \subfloat[]{
      \includegraphics[width=0.23\linewidth]{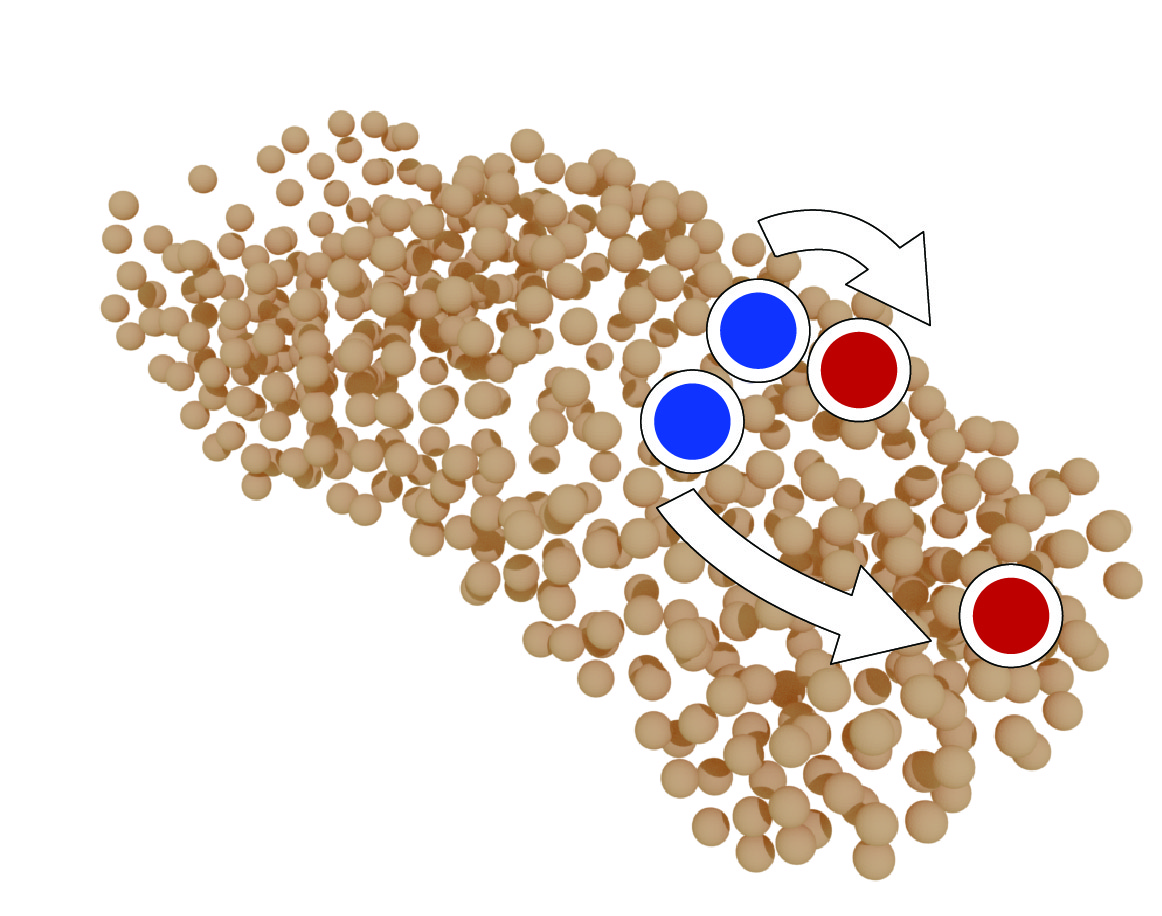}
   }
   \caption{LAP selection learned by our network is neither symmetric (a) nor invertible (b), in general; and it is not always stable (c-d), as nearby points may be mapped afar.}
   \label{fig:limit}
\end{figure}

Our paper is about learning point features, which is an essential step in most point cloud processing tasks including classification, segmentation, and more. Furthermore, we tackle arguably the most fundamental question in this setting, namely, how to find the neighbors of a point $P$ to best characterize its features.
With deep learning, many works propose to learn what $P$'s neighbors are
and their weights, as discussed in Section~\ref{sec:related}. Our work improves upon these
attentional learning methods and it is motivated by two key observations:
a) the learning of attention points should be task-specific; and b) one attention point may be
best as it is easier to attain semantic consistency; see Figures~\ref{fig:dap_count}.
The merit of our ideas has been validated through a comprehensive evaluation, demonstrating
that our one-LAP approach outperforms the use of multiple attention points, both by previous works (e.g.,
PointASNL~\cite{yan2020pointasnl}, PointWeb~\cite{zhao2019pointweb}), and when they are
learned under our framework.

Currently, our LAP selection is {\em neither symmetric nor invertible\/} in general. As shown in
Figure~\ref{fig:limit}, if two points $p$ and $q$ are symmetric on a shape, $LAP(p)$ and $LAP(q)$ may not be;
and $p \not= LAP(LAP(p))$. The figure also shows that the selection may become unstable, as one would
expect from the results of a data-driven optimization. However, it is worth noting that in general, we find the learned offset
vectors to be more stable than the final attention points, which are obtained by projecting back onto the point clouds.
These findings suggest that we do not yet have a mathematically or semantically precise way to reason about the best
attention point.
We speculate that unless we inject additional inductive biases, e.g., to enforce
consistency, into the design of our convolution, there may not be a universal interpretation, or ``geometric intuition'', to reliably
predict the best LAP.
Our current learning framework is completely data-driven, similar to all other works that learn more attention points: any geometric intuition would need to be built into the network design, not learned.

On the technical front, our method may not extend to other representations such as voxels. In voxel-based representation, the offset points need to be discretized into voxels which may cause discontinuity in gradients.

{\small
   \bibliographystyle{ieee_fullname}
   \bibliography{LAPNet}
}

\end{document}